\pdfoutput=1

\documentclass[11pt,a4paper]{article}

\usepackage[hyperref]{acl2021}

\usepackage{times}
\usepackage{latexsym}
\usepackage[utf8]{inputenc}
\usepackage{blindtext}
\usepackage{graphicx}
\usepackage{booktabs}
\usepackage{xparse}
\usepackage{caption}
\usepackage{subcaption}

\usepackage[T1]{fontenc}

\usepackage{microtype}
\usepackage{times}

\aclfinalcopy 
\def\aclpaperid{17} 

\usepackage{xspace}
\usepackage{booktabs}
\usepackage{graphicx}
\usepackage{enumitem}
\usepackage{float}
\usepackage{tikz-dependency}
\usepackage{multirow}
\usepackage{csquotes}
\usepackage{amsmath}
\usepackage{amssymb}
\usepackage{subcaption}
\usepackage{float}

\setlist[description]{leftmargin=\parindent,labelindent=0pt}

\usepackage{todonotes}
\definecolor{stefan}{rgb}{0.635,0.998,0.722}
\definecolor{jonas}{rgb}{0.998,0.722,0.635}
\definecolor{anne}{rgb}{0.68, 0.89, 0.9}

\newcommand{\sref}[1]{Sec.~\ref{sec:#1}}

\newcommand{\tref}[1]{Table~\ref{tab:#1}}
\newcommand{\fref}[1]{Figure~\ref{fig:#1}}

\renewcommand{\vec}[1]{\mathbf{#1}}

\DeclareSymbolFont{extraup}{U}{zavm}{m}{n}
\DeclareMathSymbol{\varclub}{\mathalpha}{extraup}{84}
\DeclareMathSymbol{\varspade}{\mathalpha}{extraup}{81}
\DeclareMathSymbol{\vardiamond}{\mathalpha}{extraup}{87}

\usepackage{makecell}

\title{Applying Occam's Razor to Transformer-Based Dependency Parsing: What Works, What Doesn't, and What is Really Necessary}

\author{Stefan Grünewald$^{\bigstar\vardiamond}$ \hspace*{1.5cm} Annemarie Friedrich$^\vardiamond$ \hspace*{1.5cm} Jonas Kuhn$^\bigstar$\\
    $^\bigstar$Institut für Maschinelle Sprachverarbeitung, University of Stuttgart\\
    $^\vardiamond$Bosch Center for Artificial Intelligence, Renningen, Germany\\
    \texttt{stefan.gruenewald|annemarie.friedrich@de.bosch.com}\\
    \texttt{jonas.kuhn@ims.uni-stuttgart.de}
}

\date{\today}

\NewDocumentCommand{\rot}{O{45} O{1em} m}{\makebox[#2][l]{\rotatebox{#1}{#3}}}%
\begin{document}
\maketitle

\begin{abstract}
The introduction of pre-trained transformer-based contextualized word embeddings has led to considerable improvements in the accuracy of graph-based parsers for frameworks such as Universal Dependencies (UD).
However, previous works differ in various dimensions, including their choice of pre-trained language models and whether they use LSTM layers.
With the aims of disentangling the effects of these choices and identifying a simple yet widely applicable architecture, we introduce STEPS, a new modular graph-based dependency parser.
Using STEPS, we perform a series of analyses on the UD corpora of a diverse set of languages.
We find that the choice of pre-trained embeddings has by far the greatest impact on parser performance and identify XLM-R as a robust choice across the languages in our study.
Adding LSTM layers provides no benefits when using transformer-based embeddings.
A multi-task training setup outputting additional UD features may contort results.
Taking these insights together,
we propose a simple but widely applicable parser architecture and configuration, achieving new state-of-the-art results (in terms of LAS) for 10 out of 12 diverse languages.\footnote{We release our code and pre-trained models on\\
\url{github.com/boschresearch/steps-parser}.}
\end{abstract}

\section{Introduction}
\label{sec:intro}
Recent years have seen considerable improvements in the performance of syntactic dependency parsers for frameworks such as Universal Dependencies \citep[UD;][]{de-marneffe-etal-2014-universal}.
For graph-based parsers, these improvements can in large part be attributed to two developments: (1) the introduction of deep biaffine classifiers \cite{dozat2017deep}, which now constitute the de-facto standard approach for graph-based dependency parsing, and (2) the rise of pre-trained distributed word representations, particularly transformer-based contextualized embeddings such as BERT \cite{devlin-etal-2019-bert} or RoBERTa \cite{liu2019roberta}.
Both characteristics are present in recent top-performing systems \citep{che-etal-2018-towards,straka2019evaluating,kondratyuk2019languages,kanerva-etal-2018-turku,kanerva-etal-2020-turku,che-etal-2018-towards}.

\begin{figure}
    \centering
    \includegraphics[scale=0.5]{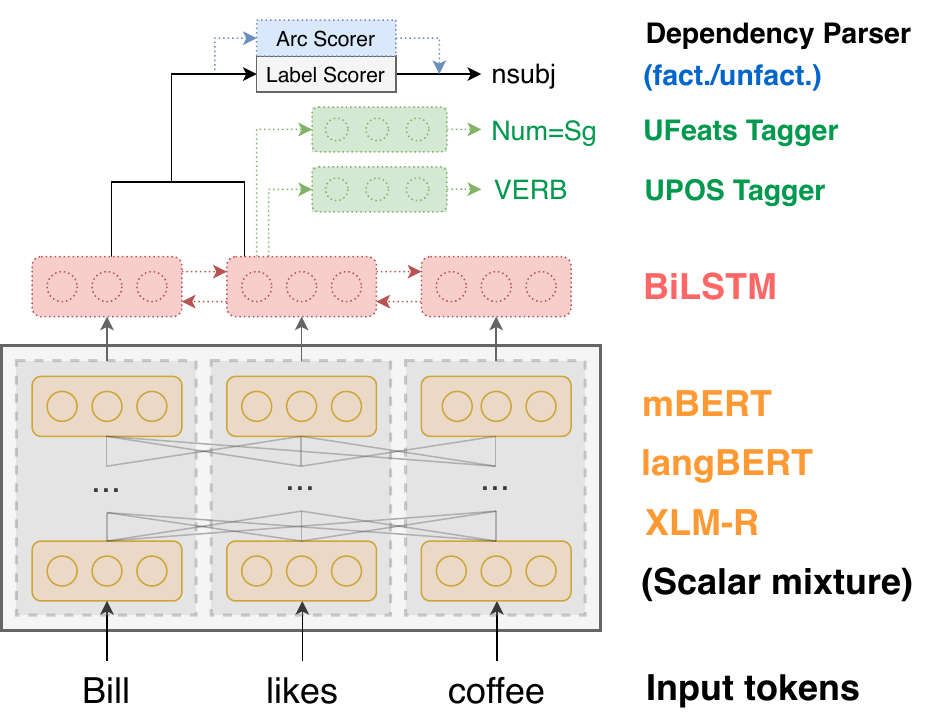}
    \caption{Modular architecture of the STEPS parser. Dotted lines denote optional components.}
    \label{fig:modular_parser}
\end{figure}

However, there remain a considerable number of implementation and configuration choices whose impact on parser performance is less well understood.
This is evidenced by the many different model configurations (see \tref{parser_configs}) present in parsers that have achieved top results in  recent shared tasks addressing UD parsing \citep{zeman-etal-2017-conll,zeman-etal-2018-conll,bouma-etal-2020-iwpt}.
The choices include (a) the particular pre-trained word embeddings or language model to use, 
(b) whether to utilize an LSTM in addition to (fine-tuned) contextualized word embeddings;
and (c) whether to use a multi-task training setup simultaneously predicting additional UD features (such as morphology or parts of speech) during parsing.

The aim of this paper is to disentangle the effects of the above factors and determine their impact on parser performance.
We appeal to the concept of Occam's razor by ways of avoiding architectural elements that do not bring about a testable advantage.
With this idea in mind, we introduce STEPS (the \textbf{S}tuttgart \textbf{T}ransformer-based \textbf{E}xtensible \textbf{P}arsing \textbf{S}ystem), a modular graph-based dependency parser which implements commonly used modules such as biaffine scorers \citep{dozat-etal-2017-stanfords,kondratyuk2019languages} or LSTM layers \citep{straka-2018-udpipe} (see \fref{modular_parser}).
Using STEPS, we perform a series of experiments on the UD treebanks of a diverse set of languages.
Our setup facilitates estimating the impact of the various architectures and configuration decisions in a comparable way.

Our most important insight is that a relatively simple architecture using biaffine heads on top of fine-tuned XLM-R \citep{conneau2019unsupervised} leads to the highest parsing accuracy for almost all languages in our study, outperforming prior systems on most languages.
Our analysis indicates that LSTM layers do not lead to benefits.
Simplifying the architecture even further by using a single scorer for edge and label prediction results in similar performance but on average leads to longer training times.
Our contributions are as follows:
\begin{enumerate}[label={(\arabic{enumi})},leftmargin=*]
    \item We introduce STEPS, a new implementation of a graph-based dependency parser
    designed to be modular and easily extensible.
    STEPS achieves new state-of-the-art UD parsing performance (in terms of LAS) for 10 out of the 12 typologically diverse languages in our study.
     We will make our code and pre-trained models for 12 languages publicly available. 
    \item We conduct a detailed experimental study, identifying components of parser architecture that are really necessary to obtain a strongly performing system that is applicable across a wide range of languages.
    The final system uses XLM-R, no LSTM layer, and a factorized edge and label scoring architecture.
    \item We show that multi-task setups predicting additional features as commonly employed in UD parsing may confound results for parsing for individual languages; we hence propose to compare parsing accuracy in unified evaluation settings in future work.
	\item We show that our parser can be easily adapted to Enhanced UD parsing, also resulting in state-of-the-art performance(in terms of ELAS) for 5 out of 7 evaluated languages.
\end{enumerate}

This paper is structured as follows.
\sref{relwork} gives the necessary background on relevant state-of-the-art neural graph-based dependency parsers as well as related work on analysing and comparing parsers.
\sref{models} describes the architecture and configuration options for our new STEPS parser, at the same time introducing the various factors studied in our experiments (\sref{evaluation}).
\sref{enhanced} presents the adaption of our system to Enhanced UD.
Finally, we discuss implications for parser choice and future parser design (\sref{conclusion}).

\section{Related Work}
\label{sec:relwork}
This section provides a brief outline of the use of contextualized word embeddings in syntactic parsers, recently developed graph-based dependency parsers, and related work on dependency parser analysis.

\begin{table}
	\centering
	\footnotesize
    \begin{tabular}{llll}
        \toprule
        \textbf{Parser} & \textbf{Pre-trained embed.} & \textbf{LSTM} & \textbf{MTL}\\
        \midrule
        StanfordNLP & word2vec, fastText & yes & no\\
        UDPipe 2.0 & word2vec, fastText & yes & yes\\
        HIT-SCIR & fastText, ELMo & yes & no\\
        UDify & mBERT & no & yes\\
        Trankit & XLM-R & no & yes\\
        \bottomrule
    \end{tabular}
    \caption{Settings for a number of previously state-of-the-art graph-based dependency parsers. \enquote{LSTM} states whether the parser makes use of an LSTM, and \enquote{MTL} states whether the parser is also trained to simultaneously predict other UD properties such as POS tags or morphological features.}
    \label{tab:parser_configs}
\end{table}

\paragraph{Contextualized Word Embeddings in Dependency Parsing.}
Like in other sub-fields of natural language processing, using contextualized word embeddings has become the de-facto standard when building syntactic parsers.
\citet{dyer-etal-2015-transition} use LSTM-based contextual representations for the stack and buffer in transition-based parsing, while \citet{kiperwasser2016simple} use BiLSTM-based feature representations for individual tokens in both graph-based and transition-based parsing.
In both of these cases, the underlying LSTM is trained simultaneously with the target task.
In contrast, recently the predominant approach towards contextualized word representations has been to pre-train systems on large-scale language modeling objectives, then taking their representations as input for a target task, optionally while continuing to fine-tune them.
This approach was initially proposed using an LSTM-based system \citep[ELMo;][]{peters-etal-2018-deep} and has since been transferred to transformers \citep[e.g., BERT;][]{devlin-etal-2019-bert}.
Transformer-based pre-trained language models have proven wildly successful and have become a standard method for a wide range of NLP tasks, including syntactic dependency parsing.

\paragraph{Recent Graph-based Parsers.}
\tref{parser_configs} shows the configurations of three parsers that were among the best-performing systems in the CoNLL 2017 and CoNLL 2018 Shared Tasks on UD parsing, as well as the more recent UDify and Trankit parsers.

\textbf{StanfordNLP} \cite{dozat-etal-2017-stanfords} was one of the first systems to apply the biaffine graph-based parser architecture to Universal Dependency parsing.
Its token representations make use of pre-trained word2vec \cite{word2vec} embeddings that are contextualized using a BiLSTM.
\textbf{UDPipe 2.0} \cite{straka-2018-udpipe} uses a multi-task setup in which POS and feature tagging, lemmatization, and dependency parsing share layers.
The system was later extended \citep[][henceforth \textbf{UDPipe+}]{straka2019evaluating} by incorporating multilingual BERT (mBERT; \citealp{devlin-etal-2019-bert}) in its token representations.
\textbf{HIT-SCIR} \cite{che-etal-2018-towards} was one of the first UD parsers to make use of contextualized pre-trained word embeddings \citep[in the form of ELMo;][]{peters-etal-2018-deep}.
The model does not make use of a multi-task training setup.
\textbf{UDify} \cite{kondratyuk2019languages} differs from previous UD parsers in two ways.
First, it does not use an LSTM layer for token representation, instead using a learned scalar mixture of mBERT layers and fine-tuning mBERT during training.
This is in contrast to the three aforementioned parsers, which do not fine-tune their pre-trained token embeddings.
Second, UDify learns a single model for all languages, concatenating all UD 2.5 training sets.
\textbf{Trankit} \cite{nguyen-etal-2021-trankit} is a recently released end-to-end UD parsing system built on the XLM-R language model. In contrast to UDify and our own STEPS parser, it does not fine-tune the entire language model, but instead inserts Adapter layers \cite{pfeiffer-etal-2020-adapterhub,pfeiffer-etal-2020-mad} to efficiently create language-specific models for 56 languages.

\paragraph{Multi-Purpose Parsers.}
Other parsers with modular or extensible architectures include \textbf{Alto} \citep{gontrum-etal-2017-alto}, a prototyping tool for new grammar formalisms based on Interpreted Regular Tree Grammars (IRTGs), and \textbf{PanParser} \citep{aufrant2018panparser}, a modular framework for transition-based dependency parsing.
In contrast to these two, STEPS is a graph-based dependency parser that focuses on easy configuration of different transformer-based language models and neural architecture variants.

\paragraph{Parser Analyses and Comparisons.}
\label{sec:relwork_analysis}
Recent years have seen a wide range of studies comparing different language models for dependency parsing (e.g., \citealp{kanerva-etal-2018-turku, pyysalo2020wikibert, smith-etal-2018-investigation}).
Additionally, several studies have investigated the amount of implicit syntactic information captured in pre-trained LMs such as ELMo and BERT \cite{tenney-etal-2019-bert, tenney2019you, hewitt-manning-2019-structural}.
Conversely, several studies have investigated the utility of structural features 
for dependency parsing in the presence of LSTMs and/or contextualized word embeddings, generally finding that their impact is diminished in the presence of contextual information \cite{falenska-kuhn-2019-nonutility, fonseca-martins-2020-revisiting}.

\citet{kulmizev-etal-2019-deep} compare the effect of deep contextualized word embeddings on transition-based and graph-based dependency parsers, showing that their inclusion makes the two approaches virtually equivalent in terms of parsing accuracy.
Our work is similar to theirs in the sense that we also evaluate several very different dimensions of parser architecture at the same time, utilizing the same underlying backbone and thus ensuring comparability across experiments.

\section{STEPS: A Modular Graph-Based Dependency Parser}
\label{sec:models}

In this section, we describe our modular dependency parser STEPS (\textbf{S}tuttgart \textbf{T}ransformer-based \textbf{E}xtensible \textbf{P}arsing \textbf{S}ystem).
Each subsection focuses on a particular aspect of the parser setup, providing background on its usage and its potential impact on parser performance.

\subsection{Input Token Representation}
\label{sec:input_representation}
STEPS provides a number of different options for input token representation.
As \tref{parser_configs} shows, parsers have made use of a variety of pre-trained embeddings, with transformer-based language models having become the predominant current approach.
We hence focus on the latter and compare multilingual BERT \citep[\textbf{mBERT};][]{devlin-etal-2019-bert}, language-specific BERTs (\textbf{langBERT}), and the multilingual \textbf{XLM-R-large} model \cite{conneau2019unsupervised}.
XLM-R utilizes the pre-training optimizations first proposed for RoBERTa \citep{liu2019roberta}, which includes training on a considerably larger amount of data.
A detailed overview of all transformer models used in our experiments is provided in the second column of \tref{lang_data_models}.

\begin{table*}
    \centering
    \footnotesize
    \begin{tabular}{lll}
    \toprule
     \textbf{Language} & \textbf{Transformer LM} & \textbf{UD Treebank}\\
    \midrule
        Arabic & ArabicBERT-large \cite{safaya2020kuisail} & PADT \cite{smrz2008prague}\\
        Chinese & Chinese BERT \cite{devlin-etal-2019-bert} & GSD \\
        Czech & Slavic-BERT \cite{arkhipov-etal-2019-tuning} & PDT \cite{bejcek-etal-2012-prague}\\
        English & RoBERTa-large \cite{liu2019roberta} & EWT \cite{silveira14gold}\\
        Finnish & FinBERT \cite{virtanen2019multilingual} & TDT \\
        German & German BERT (\url{github.com/dbmdz/berts}) & GSD \cite{mcdonald-etal-2013-universal}\\
        Hindi & WikiBERT-Hindi \cite{pyysalo2020wikibert} & HDTB \cite{bhathindi,palmer2009hindi}\\
        Italian & Italian BERT-XXL (\url{github.com/dbmdz/berts}) & ISDT\\
        Japanese & WikiBERT-Japanese \cite{pyysalo2020wikibert} & GSD\\
        Korean & KR-BERT \cite{lee2020kr} & Kaist \cite{chun-etal-2018-building}\\
        Latvian & WikiBERT-Latvian \cite{pyysalo2020wikibert} & LVTB\\
        Russian & RuBERT \cite{kuratov2019adaptation} & SynTagRus \cite{droganova2018data}\\
    \midrule
        Multilingual & mBERT \cite{devlin-etal-2019-bert} & --\\
        Multilingual & XLM-R \cite{conneau2019unsupervised} & --\\
    \bottomrule
    \end{tabular}
    \caption{Language models and UD treebanks used in our experiments. Citations for treebanks are given where provided in treebank repository documentation.}
    \label{tab:lang_data_models}
\end{table*}

STEPS represents each token $i$ using a vector $\vec{r}_i$ corresponding to the embedding of its first word-piece token.
Following \citet{kondratyuk2019languages}, we compute token embeddings as weighted sums of the representations of the respective tokens given by the internal transformer encoder layers, resulting in either 768- oder 1024-dimensional embeddings depending on the transformer model used.
Coefficients for this sum are learned during training, and layer dropout is applied in order to prevent the model from focusing on particular layers.
Our model learns a different set of these coefficients of for each output task (see \sref{classifier} and \sref{multitask} below).
In addition to the above described \textbf{transformer-only} setting, we also compute another version of token embeddings by feeding the embeddings computed by the sum operations into a multi-layer bidirectional \textbf{LSTM} (BiLSTM), whose per-token output then constitutes $\vec{r}_i$.

\subsection{Biaffine Classifier Architecture}
\label{sec:classifier}
STEPS makes use of biaffine classifiers as proposed by \citet{dozat2017deep}, which have become the de-facto standard method for graph-based dependency parsing.
In a first step, a head representation $\vec{h}^{head}_i$ and a dependent representation $\vec{h}^{dep}_i$ are created for each input token $i$ represented as embedding vector $\vec{r}_i$ via two single-layer feedforward neural networks:
\begin{align}
	\vec{h}^{head}_i &= \text{FNN}^{head}(\vec{r}_i)\\
	\vec{h}^{dep}_i &= \text{FNN}^{dep}(\vec{r}_i)
\end{align}
These representations are then fed into the biaffine function, which maps head--dependent pairs $(i,j)$ onto vectors $\vec{s}_{i,j}$ of arbitrary size:
\begin{align}
\vec{s}_{i,j} &= \text{Biaff}\big( \vec{h}^{head}_i, \vec{h}^{dep}_j \big)\\
\hspace*{-1mm}\text{Biaff}(\vec{x}_1, \vec{x}_2) &= \vec{x}^\top_1 \vec{U} \vec{x}_2 + W(\vec{x}_1 \oplus \vec{x}_2) + \vec{b} \label{eq:biaff}
\end{align}
$\vec{U}$, $W$ and $\vec{b}$ are learned parameters; $\oplus$ denotes the concatenation operation.
The scores $s_{i,j}$ can now be leveraged in different ways to construct an output tree or graph; this will be described next.

First, the \textbf{factorized} approach \citep{dozat2017deep} uses two instances of biaffine classifiers.
The first classifier (the \enquote{arc scorer}) is responsible for predicting which (unlabeled) edges exist in the output structure.
It predicts, for each token, a probability distribution over potential syntactic heads (i.e., all other tokens in the sentence).
We then feed the log-probabilities to the Chu-Liu/Edmonds maximum spanning tree algorithm \cite{chuliu1965shortest, edmonds1967optimum} and label the resulting tree using the label scorer.
The second classifier (the \enquote{label scorer}) then assigns dependency labels to edges predicted in the first step.

The \textbf{unfactorized} approach, proposed by \citet{dozat-manning-2018-simpler} for semantic graph parsing, uses only a single biaffine classifier (namely the label scorer).
Non-existence of dependencies is encoded using simply another label ($\varnothing$).
We adapt this approach to tree parsing by discarding the arc scorer and computing the edge weights for the Chu-Liu/Edmonds MST algorithm as $\log (1 - P(\varnothing))$ in order to extract a labeled dependency tree directly.
To the best of our knowledge, this is the first time that the unfactorized architecture has been applied to the parsing of dependency tree structures.

\subsection{Multi-Task Training}
\label{sec:multitask}
We study the effects of a multi-task training setup by implementing two approaches to training our parser: (a) \textbf{dep-only}, in which the model is trained only on syntactic dependencies; and (b) multi-task learning (\textbf{MTL}), in which the model additionally predicts universal part-of-speech tags (UPOS) and morphological features (UFeats).
We follow \citet{kondratyuk2019languages} by learning different coefficients for the transformer layers for these tagging tasks (see \sref{input_representation}) and then using a single-layer feed-forward neural network to extract logit vectors over the respective label vocabularies.
By default, the loss for the entire system is computed as the sum of losses for the individual output modules (UPOS tagger, UFeats tagger, and dependency parser).
However, we also add the option of scaling the loss of the individual output modules in order to prevent individual tasks from overwhelming the system as a whole (see \sref{mtl_eval}).

\section{Experiments}
\label{sec:evaluation}

This section describes our experimental setup and reports the results of our experiments on pre-trained embeddings (\sref{experiments-embeddings}), factorized vs. unfactorized parser architecture (\sref{experiments:factorization}), LSTM usage (\sref{experiments:lstm}), and multi-task training (\sref{mtl_eval}).

\subsection{Experimental Setup}

\paragraph{Languages and treebanks.}
We select 12 languages, covering a diverse range of language families and writing systems, by applying linguistic criteria similar to those outlined by \citet{delhoneux-etal-2017-oldvsnew}.
For each language, we select the largest available treebank from UD 2.6 for which token data is freely available.
These treebanks are listed in the third column of \tref{lang_data_models}.
In all of our experiments, we use gold tokens and train language-specific models, testing on the test set of the respective treebank.

\paragraph{Evaluation metrics.}
We compute UAS and LAS using the official evaluation script for the CoNLL 2018 Shared Task.\footnote{\url{https://universaldependencies.org/conll18/conll18_ud_eval.py}}
UAS (Unlabeled Attachment Score) computes the fraction of tokens that have been assigned the correct syntactic head.
LAS (Labeled Attachment Score) records the fraction of tokens that have been assigned the correct syntactic head with the correct edge label.

\subsection{Implementation}
Our parser is implemented in Python, using PyTorch \cite{paszke2019pytorch} and the Huggingface Transformers library \cite{wolf2019huggingface}.
Training is performed on a single nVidia Tesla V100 GPU.

\paragraph{Hyperparameters.}
We aim to obtain a simple yet high-performing hyperparameter configuration.
To do so, we start out with the configuration of UDify, which is architecturally quite similar to STEPS, and tune parameters using grid search in ca. 40 runs on a small development set (consisting of English, Arabic, and Korean data), aiming at a simplified setup that achieves good results across these diverse languages.
The hyperparameters examined by us were
\begin{itemize}
    \item Hidden size of the biaffine classifier (256 / 512 / 768 / 1024)
    \item Batch size (16 / 32)
    \item Base learning rate ($7e^{-6}$ to $5e^{-5}$)
    \item Early stopping patience (10 / 15 / 20 epochs)
    \item Learning rate schedule (constant LR / warmup only / cosine annealing / Noam)
\end{itemize}

In large part, our final settings are identical to UDify's values with the following differences.
We use the AdamW optimizer \cite{loshchilov2018decoupled} instead of Adam; we perform neither label smoothing nor gradient clipping; and we do not use differential learning rates.
In addition, we do not train for a fixed number of epochs, but instead stop once performance on the validation set does not increase for 15 epochs, or after at most 24 hours.

For model variants involving LSTMs, we tuned the hyperparameters involved in these layers (number of layers; hidden size; dropout; learning rate) in a second round of optimization consisting of 15 trials of random search on the English data.
We then picked the two best-performing models and ran them on the other languages, finding that one of them performed best on all languages.

All of our final hyperparameter settings can be found in \tref{hyperparameters}.

\begin{table}
	\centering
	\footnotesize
	\begin{tabular}{ll}
		\toprule
		\multicolumn{2}{c}{\textbf{Transformer LM}} \\
		Token mask probability & 0.15\\
		Layer dropout & 0.1\\
		Hidden dropout & 0.2\\
		Attention dropout & 0.2\\
		Output dropout & 0.5\\
		\multicolumn{2}{c}{\textbf{Biaffine classifier}}\\
		Arc scorer dimension & 768 or 1024\textsuperscript{a}\\
		Label scorer dimension & 256 or 768/1024\textsuperscript{b}\\
		Dropout & 0.33\\
		\multicolumn{2}{c}{\textbf{LSTM}}\\
		Hidden size & 330\\
		Number of layers & 3\\
		Dropout & 0.5\\
		LSTM learning rate & $5e^{-4}$\\
		\multicolumn{2}{c}{\textbf{Optimization}}\\
		Optimizer & AdamW\\
		$\beta_1$, $\beta_2$ & 0.9, 0.999\\
		Weight decay & 0\\
		Batch size & 32\\
		Base learning rate & $4e^{-5}$\\
		LR schedule & Noam\\
		LR warmup & 1 epoch\\
		\bottomrule
	\end{tabular}
	\caption{Hyperparameter values. \textsuperscript{a}Identical to hidden size of the transformer encoder. \textsuperscript{b}256 in factorized models, hidden size of transformer encoder in unfactorized models.}
	\label{tab:hyperparameters}
\end{table}

\begin{table*}
	\centering
	\footnotesize
	\setlength{\tabcolsep}{5pt}
	\begin{tabular}{lcccccccccccc}
		\toprule
		& ar   & cs  & de  & en  & fi & hi   & it   & ja  & ko    & lv   & ru  & zh \\
		& PADT & PDT & GSD & EWT & TDT & HDTB & ISDT & GSD & Kaist & LVTB & STR & GSD\\
		\midrule
		UDPipe+ & 84.62 & 92.56 & 84.06 & 90.40 & 89.49 & 92.50 & 93.38 & \textbf{94.27} & 87.54 & 84.50 & 93.68 & 86.74 \\
		UDify  & 82.88 & 92.88 & 83.59 & 88.50 & 82.03 & 91.46 & 93.69 & 92.08 & 84.52 & 85.09 & 93.13 & 83.75 \\
		UDify\textsubscript{mono} & 83.34 & 91.58 & 84.28 & 89.52 & 86.74 & 91.44 & 93.14 & 92.14 & 86.45 & 85.45 & 92.32 & 82.95\\
		Trankit\textsubscript{large} & 86.51 & 93.11 & \textbf{86.27} & 91.64 & 94.31 & 93.17 & 94.63 & 78.14 & 40.76 & 91.76 & 95.16 & 87.38\\
		\midrule
		STEPS\textsubscript{mBERT}      & 83.80 & 92.69 & 84.08 & 89.16 & 88.91 & 91.30 & 93.13 & 92.22 & 24.49 & 85.05 & 93.74 & 84.94 \\ 
		STEPS\textsubscript{langBERT}   & \textbf{86.60} & 92.99 & 85.87 & \textbf{91.98} & 93.57 & 91.16 & 94.25 & 92.98 & 84.56 & 82.61 & 94.44 & 86.20 \\
        STEPS\textsubscript{XLM-R}  & 86.55 & \textbf{94.58} & 86.07 & 91.91 & \textbf{94.36} & \textbf{93.34} & \textbf{94.86} & 94.10 & \textbf{89.93} & \textbf{91.93} & \textbf{95.30} & \textbf{87.75} \\
        \midrule
        STEPS\textsubscript{XLM-R-LSTM}   & 86.41 & 94.52 & 86.20 & 91.58 & 93.92 & 93.28 & 94.57 & 94.01 & 89.91 & 91.61 & 95.27 & 86.96 \\
		STEPS\textsubscript{XLM-R-unfact} & 86.32 & 94.38 & 86.19 & 91.57 & 94.11 & 93.21 & 94.58 & 93.58 & 89.86 & 91.76 & 95.17 & 87.47\\
		\bottomrule
	\end{tabular}	
	\caption{Labeled Attachment Score (LAS) for \textbf{basic dependency parsing} varying input embeddings and architecture. STEPS scores are averages of three runs.} 
	\label{tab:basic_deps}
\end{table*}

\subsection{Impact of Pre-Trained Word Embeddings}
\label{sec:experiments-embeddings}
We first evaluate how parsing performance differs when varying the underlying pre-trained language model.
We here do not include an LSTM layer and perform only dependency parsing.
\tref{basic_deps} shows results for all 12 treebanks used in this study.
UDPipe+ refers to the version of UDPipe enhanced with BERT and Flair embeddings proposed by \citet{straka2019evaluating} and described in \sref{relwork_analysis}.
UDify refers to the original system trained on all UD languages without treebank-specific fine-tuning.
As multilingual training usually results in improved performance for low-resource languages at the cost of lowering scores for high-resource languages \cite{ustun-etal-2020-udapter}, for meaningful comparison, we train UDify\textsubscript{mono} on single treebanks.
Trankit\textsubscript{large} refers to the version of Trankit which uses XLM-R-large as the underlying language model, same as STEPS\textsubscript{XLM-R}.

STEPS\textsubscript{mBERT} roughly corresponds to UDify\textsubscript{mono}, and indeed the models overall perform similarly.
We attribute differences to slightly different training setups.
While UDify is trained for 80 epochs, STEPS employs early stopping after 15 epochs without improvement.
Moreover, we did not disable multi-task learning for parallel UD feature prediction in UDify\textsubscript{mono}, and this may be an explanation why STEPS\textsubscript{mBERT} does much better on Finnish, Czech and Russian, where morphological features may be harder to predict.
(For a principled comparison of multi-task setups, see \sref{mtl_eval}.)
By contrast, UDPipe+ often outperforms UDify, UDify\textsubscript{mono}, and STEPS\textsubscript{mBERT}, which is likely due to the fact that it trains its own word embeddings in addition to mBERT and additionally makes use of character-level representations via GRUs.

Parsing accuracy of STEPS is very high across the board, with new state-of-the-art results being achieved on all languages except Japanese and German.
For most languages, the best results are achieved using STEPS\textsubscript{XLM-R}, with STEPS\textsubscript{langBERT} coming in second.
In contrast, using mBERT is not the best option on any treebank.
In fact, the only languages for which mBERT achieves better results than langBERT in our experiments are Latvian and Hindi.\footnote{In a similar study comparing mBERT- and langBERT-based parsers, \citet{kanerva-etal-2020-turku} also found Latvian to be one of the few languages for which mBERT outperformed the language-specific (WikiBERT) version. Both the Latvian and the Hindi Wikipedias are rather small, consisting of only 21M and 35M tokens, respectively \citep{pyysalo2020wikibert}.}
While using langBERT usually yields worse parsing accuracy than XLM-R, results are roughly on par for Arabic and English. 
We note that the language-specific models we chose for these treebanks (ArabicBERT-large and RoBERTa-large, respectively) are the only ones with a number of trainable parameters similar to XLM-R,
while all others have a considerably smaller number of parameters.
This highlights the importance of model size in pre-trained word embeddings.

STEPS\textsubscript{XLM-R} and Trankit\textsubscript{large} show rather similar performance overall, which is to be expected given the fact that both are built on the same underlying language model (XLM-R-large).
The slight advantage for STEPS\textsubscript{XLM-R} observed on most languages may stem from the fact that it fine-tunes the entire transformer model instead of merely adding Adapter layers, and that it does not use a multi-task training setup (cf. \sref{mtl_eval}).
Interestingly, on Finnish and Latvian, both systems outperform other existing parsers by very large margins (around 4.9 and 6.5 LAS, respectively).
We assume that there are two main reasons for this.
First, XLM-R is pre-trained on CommonCrawl data \cite{conneau2019unsupervised, wenzek-etal-2020-ccnet} as opposed to Wikipedia dumps, which results not only in several orders of magnitude more training data (over 1 billion tokens for both languages), but also presumably more heterogenous data, which may provide better generalizations for the domains in our test data.\footnote{Both fi-TDT and lv-LVTB contain, among others, \enquote{non-standard} data such as blog entries, legal texts, and spoken language \cite{haverinen2014building, pretkalnicna2018deriving}. }
Second, XLM-R has a much larger vocabulary size than mBERT (250k vs. 100k), 
which means it may account better for the rich morphology of these languages. On average, a Finnish (Latvian) token is split up into 2.4 (2.1) word pieces when using mBERT, but only 1.9 (1.8) word pieces when using XLM-R.

Finally, we note that STEPS\textsubscript{mBERT} and Trankit\textsubscript{large} perform extremely poorly on Korean (24.49/40.76 LAS on average), indicating that the models do not properly learn from the data.
We assume that this may be a tokenization or character encoding issue unique to the Korean-Kaist treebank.\footnote{As pointed out by an anonymous reviewer, Korean-Kaist uses a rather different tokenization strategy than other UD treebanks, with tokens corresponding to larger chunks. Relying on just the first word pieces for token embeddings may be problematic in this context.}
However, a similar pattern is not observed for any of the other parser models, and we were unfortunately unable to identify the exact cause despite our best efforts.

\subsection{Impact of LSTM Layer}
\label{sec:experiments:lstm}
We evaluate the performance of a system identical to STEPS\textsubscript{XLM-R} described above, but with 3 additional BiLSTM layers added on top of the language model (STEPS\textsubscript{XLM-R-LSTM} in \tref{basic_deps}).
Changes in performance are generally small.
With the exception of German, including LSTM layers actually decreases parsing accuracy slightly.
The LSTM model contains more trainable parameters and also makes use of differential learning rates, yet, we did not find any meaningful differences in convergence speed and training times.
Hence, we conclude that when fine-tuning an underlying transformer-based language model, adding LSTM layers on top is not necessary.
However, results may differ for systems that additionally train their own token embeddings or make use of character-based representations, both of which we do not address in our experiments.

\subsection{Impact of Factorization}
\label{sec:experiments:factorization}
\citet{dozat-manning-2018-simpler} show that for semantic dependency graph parsing, a simplified parser architecture predicting edge presence and edge labels from the same scoring matrix achieves largely identical results compared to a model using two separate classifiers for arcs and labels.
We here dive into the question whether such an unfactorized approach is also able to achieve competitive results in syntactic tree parsing.
We do so by implementing a version of STEPS\textsubscript{XLM-R} that makes use of the unfactorized approach as descibed in \sref{classifier}.

Results of our experiments can be found in the row labeled STEPS\textsubscript{XLM-R-unfact} in \tref{basic_deps}.
Overall, performance of the unfactorized approach is very close to the factorized version, but slightly lower.
While this shows that the unfactorized approach is indeed viable for tree parsing, analysis of the training times reveals an increase by ca. 30\,\% on average when using the unfactorized model, indicating that the shared scorer takes a longer time to converge. 

In light of these results, we propose to stick with the factorized version for syntactic tree parsing.
At least in a research setting, shorter training times allow for a larger set of experiments and thus ultimately in using fewer resources.
When applying the parser, differences in model size and parsing time are negligible.

\subsection{Impact of Multi-Task Approach}
\label{sec:mtl_eval}

\begin{table*}
	\centering
	\footnotesize
	\setlength\tabcolsep{2pt} 
	\begin{tabular}{llcccc}
		\toprule
		\textsc{Treebank} & \textsc{Model} & UPOS & \textsc{UFeats} & UAS & LAS \\
		\midrule
		 & Trankit\textsubscript{large} & 95.47 & \textbf{95.54} & 90.90 & 86.51\\
		Arabic      & STEPS\textsubscript{dep-only} & -- & -- & \textbf{90.96} & \textbf{86.55} \\
        (ar\_padt)  & STEPS\textsubscript{MTL} & \textbf{97.24} & 94.89 & 90.34 & 86.01 \\
		           & STEPS\textsubscript{MTLscale} & 96.47 & 87.74 & 90.80 & 86.41 \\
		\midrule
		& Trankit\textsubscript{large} & 99.37 & \textbf{98.23} & 95.51 & 93.11\\
        Czech PDT  & STEPS\textsubscript{dep-only} & -- & -- & \textbf{95.89} & \textbf{94.58} \\
        (cs\_pdt)  & STEPS\textsubscript{MTL} & \textbf{99.41} & 98.06 & 95.59 & 94.19 \\
		           & STEPS\textsubscript{MTLscale} & 98.97 & 94.20 & 95.85 & 94.52 \\
		\midrule
		& Trankit\textsubscript{large} & \textbf{95.48} & \textbf{91.91} & \textbf{90.11} & \textbf{86.27}\\
        German     & STEPS\textsubscript{dep-only} & -- & -- & 90.02 & 86.07 \\
        (de\_gsd)  & STEPS\textsubscript{MTL} & 95.40 & \textbf{91.91} & 89.53 & 85.46 \\
		           & STEPS\textsubscript{MTLscale} & 94.65 & 83.33 & 89.74 & 85.80 \\
		\midrule
		& Trankit\textsubscript{large} & \textbf{97.91} & \textbf{98.04} & 93.59 & 91.64\\
	    English       & STEPS\textsubscript{dep-only} & -- & -- & \textbf{93.90} & \textbf{91.91} \\
        (en\_ewt)     & STEPS\textsubscript{MTL} & 97.84 & 98.02 & 93.47 & 91.50 \\
		              & STEPS\textsubscript{MTLscale} & 96.58 & 96.49 & 93.80 & 91.78 \\
		\midrule
		& Trankit\textsubscript{large} & \textbf{98.72} & \textbf{97.07} & 95.55 & 94.31\\
		Finnish TDT    & STEPS\textsubscript{dep-only} & -- & -- & \textbf{95.69} & \textbf{94.36} \\
        (fi\_tdt)      & STEPS\textsubscript{MTL} & 98.52 & 96.75 & 94.62 & 93.11 \\
		               & STEPS\textsubscript{MTLscale} & 98.19 & 88.70 & 95.59 & 94.26 \\
		\midrule
		& Trankit\textsubscript{large} & \textbf{98.12} & 93.98 & 96.16 & 93.17\\
        Hindi       & STEPS\textsubscript{dep-only} & -- & -- & 96.11 & 93.34 \\
        (hi\_hdtb)  & STEPS\textsubscript{MTL} & 98.09 & \textbf{94.49} & 95.96 & 93.03 \\
		            & STEPS\textsubscript{MTLscale} & 97.51 & 88.60 & \textbf{96.18} & \textbf{93.39} \\
		        		\bottomrule
	\end{tabular}   
\quad
	\begin{tabular}{llcccc}
		\toprule
		\textsc{Treebank} & \textsc{Model} & UPOS & \textsc{UFeats} & UAS & LAS \\
		\midrule
                   & Trankit\textsubscript{large} & 98.80 & 98.43 & 95.93 & 94.63\\
        Italian     & STEPS\textsubscript{dep-only} & -- & -- & \textbf{96.25} & \textbf{94.86} \\
        (it\_isdt)  & STEPS\textsubscript{MTL} & \textbf{98.81} & \textbf{98.58} & 95.80 & 94.36 \\
		            & STEPS\textsubscript{MTLscale} & 98.43 & 94.28 & 96.09 & 94.69 \\
		\midrule
		& Trankit\textsubscript{large} & 92.57 & 97.58 & 86.62 & 78.14\\
        Japanese         & STEPS\textsubscript{dep-only} & -- & -- & \textbf{95.62} & \textbf{94.10} \\
        (ja\_gsd)        & STEPS\textsubscript{MTL} & \textbf{98.21} & \textbf{99.98} & 95.38 & 93.78 \\
		           & STEPS\textsubscript{MTLscale} & 96.94 & 99.91 & 95.53 & 94.00 \\
		\midrule
        & Trankit\textsubscript{large} & 69.86 & 98.95 & 67.96 & 40.76 \\
        Korean      & STEPS\textsubscript{dep-only} & -- & -- & \textbf{91.71} & \textbf{89.93}
 \\
        (ko\_kaist) & STEPS\textsubscript{MTL} & \textbf{96.41} & \textbf{100.00} & 91.48 & 89.73 \\
		            & STEPS\textsubscript{MTLscale} & 93.90 & \textbf{100.00} & 91.53 & 89.77 \\
		\midrule
        & Trankit\textsubscript{large} & \textbf{97.83} & \textbf{95.38} & 94.19 & 91.76 \\
		  Latvian         & STEPS\textsubscript{dep-only} & -- & -- & \textbf{94.32} & \textbf{91.93} \\
		 (lv\_lvtb)          & STEPS\textsubscript{MTL} & 97.73 & 94.79 & 93.44 & 90.99 \\
		           & STEPS\textsubscript{MTLscale} & 96.72 & 81.42 & 93.97 & 91.61 \\
		\midrule
		& Trankit\textsubscript{large} & \textbf{99.34} & \textbf{98.47} & 96.18 & 95.16 \\
        Russian          & STEPS\textsubscript{dep-only} & -- & -- & \textbf{96.32} & \textbf{95.30} \\
        (ru\_syntagrus)  & STEPS\textsubscript{MTL} & 99.27 & 98.34 & 95.94 & 94.88 \\
		                 & STEPS\textsubscript{MTLscale} & 98.99 & 95.91 & 96.19 & 95.20 \\
		\midrule
		& Trankit\textsubscript{large} & 96.83 & 99.48 & 90.03 & 87.38\\
        Chinese    & STEPS\textsubscript{dep-only} & -- & -- & \textbf{90.72} & \textbf{87.75} \\
        (zh\_gsd)  & STEPS\textsubscript{MTL}      & \textbf{97.20} & \textbf{99.50} & 89.70 & 86.86 \\  
		            & STEPS\textsubscript{MTLscale} & 95.21 & 98.53 & 90.31 & 87.39 \\       
		\bottomrule
	\end{tabular}
	
	\caption{Results for \textbf{basic dependency parsing} vs. \textbf{parsing and feature prediction} (\textbf{multi-task}) for STEPS\textsubscript{XLM-R}. Scorres are averages of three runs. For UPOS and UFeats, we report accuracy.}
	\label{tab:multitask_basic}
\end{table*}

Finally, we analyze how performance changes when predicting UPOS and UFeats in addition to dependencies.
For these experiments, we use XLM-R as input embeddings and a factorized architecture.
For UFeats, we follow UDify's approach and consider each possible combination of morphological features a unique label.
As shown in \tref{multitask_basic}, STEPS\textsubscript{MTL} achieves very high accuracies for UPOS and UFeats, performing on par with or only slightly worse than the previous state of the art (Trankit\textsubscript{large}) for most languages.
However, we find that compared to the dependency-only system, parsing accuracy drops considerably in the multi-task setting (up to over 1 LAS for Finnish).

During training of STEPS\textsubscript{MTL}, accuracy on the validation set increased very rapidly for the tagging tasks and reached levels close to the final values after only a few epochs, while accuracy for the parsing task increased much slower.
This suggests that the loss for the tagging tasks might overwhelm the system as a whole, causing the parser modules to underfit.
We therefore also test STEPS\textsubscript{MTLscale}, in which the loss for UPOS and UFeats is scaled down to 5\% during training.
STEPS\textsubscript{MTLscale} performs close to STEPS\textsubscript{MTL}, even outperforming it in the case of Hindi.
In turn, however, accuracy for UPOS and particularly UFeats drops considerably.

To sum up, our experiments indicate that multi-task setups as commonly employed in UD parsing have a non-negligible effect on parsing performance.
Hence, when comparing parser performance, it is crucial to take potential multi-task setups into account.
If the respective setups differ, ignoring them may result in misleading interpretations of parsing performance of model architectures (unless the variable of interest is the multi-task setup itself).

\subsection{Summary}
Our experimental findings can be summarized as follows:
(a) Choice of pre-trained embeddings has the greatest impact on parser performance, with XLM-R yielding the best results in most cases;
(b) adding LSTM layers is not necessary when working with a large fine-tuned language model;
(c) a factorized parser architecture is preferable due to faster training;
(d) when using a multi-task approach incorporating UPOS and UFeats prediction, there is a tradeoff between tagging and parsing accuracy, and conclusions regarding architecture should be drawn by comparing experiments performed in the same setting.
Crucially, one of the simplest parsers in our evaluation (STEPS\textsubscript{XLM-R}) achieves the best results overall, often surpassing more complex previous work.

\section{Enhanced UD Parsing with STEPS}
\label{sec:enhanced}

In order to determine whether our conclusions also hold for the related graph parsing task of Enhanced UD \cite{schuster-manning-2016-enhanced}, we run an additional batch of experiments on 7 treebanks from the IWPT 2020 Shared Task \citep{bouma-etal-2020-iwpt}.

\paragraph{Modifications to STEPS.}

We modify STEPS to generate dependency graphs using a factorized approach as proposed by \citet{dozat-manning-2018-simpler} for semantic dependency parsing,
weighting the losses of the edge and label scorers:
\begin{equation}\label{eq:loss_scaling}
   \ell = \lambda_{edge} \ell_{edge} + \lambda_{label} \ell_{label}\text{.}
\end{equation}
After tuning on English in a set of preliminary experiments, we set the hyperparameters $\lambda_{edge}$ to 1.0 and $\lambda_{label}$ to 0.05.
For comparison, we also evaluate the unfactorized version of our parser. 

While enhanced UD does not require output graphs to be trees, it imposes the constraint that every node must be reachable from the root.
We use the heuristic proposed by \citet{gruenewald2020robertnlp} for graph post-processing, which greedily adds the highest-scoring edge from a node that is reachable from the root to a node that is unreachable from the root until the condition is fulfilled.

Furthermore, for certain relations such as \textit{nmod} or \textit{obl}, enhanced UD allows for the inclusion of lexical material (such as prepositions) in dependency labels.
To avoid data sparsity issues resulting from the increase in the number of dependency labels, we follow the label de- and re-lexicalization strategy proposed by \citet{gruenewald2020robertnlp}, replacing lexical materials in labels with with placeholders such as \textit{obl:[case]}.
At prediction time, lexicalized parts of the labels can be retrieved from the respective child nodes in the graph.
We apply this strategy for all languages in our study except Finnish and Russian (which do not have lexicalized labels) and Arabic (for which we additionally look up lemmas of the lexical material using a simple majority baseline method).

\paragraph{Experimental Results.}

We compare our results against \textbf{TurkuNLP}, a modified version 
of UDify which scored 1st in the official evaluation of the IWPT 2020 Shared Task, and \textbf{ShanghaiTech}, which scored 1st in the unofficial post-evaluation.

We evaluate in terms of ELAS (Enhanced LAS, i.e., F1 score over the set of enhanced dependencies in the system output and the gold standard)
using the official evaluation script for the IWPT 2020 Shared Task\footnote{\url{https://universaldependencies.org/iwpt20/iwpt20_xud_eval.py}} and report per-treebank results for TurkuNLP and ShanghaiTech as submitted.\footnote{\url{https://universaldependencies.org/iwpt20/Results.html}}
To ensure comparability with previous work, we compute our results using raw text as input and using Stanza \cite{qi2020stanza} for tokenization and sentence segmentation.
\tref{enhanced_deps} reports our results.
Our parser achieves very high accuracy, outperforming TurkuNLP and ShanghaiTech on all evaluated languages except Arabic and Czech.
Notably, the latter system also uses XLM-R embeddings, but with a more complex parser architecture.

\begin{table}[t]
	\centering
	\footnotesize
	\setlength\tabcolsep{5pt} 
    \begin{tabular}{lcc|cc}
    \toprule
    & Turku- & Shanghai- & \multicolumn{2}{c}{STEPS\textsubscript{XLM-R}} \\
    & NLP & Tech & fact & unfact\\
    \midrule
    ar-PADT  & \textbf{77.83} & 77.73 & 77.42 & 77.68 \\
    cs-PDT   & 88.17 & \textbf{90.63} & 89.49 & 89.43 \\
    en-EWT   & 86.14 & 86.30 & 87.11 & \textbf{87.28} \\
    fi-TDT   & 89.24 & 89.97 & \textbf{91.53} & 91.51 \\
    it-ISDT  & 91.54 & 91.49 & 92.35 & \textbf{92.39} \\
    lv-LVTB  & 84.94 & 87.64 & \textbf{89.04} & 88.95 \\
    ru-STR   & 90.69 & 92.31 & 93.68 & \textbf{93.75} \\
    \bottomrule
    \end{tabular}
    \caption{Results (ELAS) for \textbf{enhanced dependency parsing}. Scores are averages of three runs.}
    \label{tab:enhanced_deps}
\end{table}

Unlike in tree parsing, the unfactorized system actually slightly outperforms the factorized system on a number of languages, with the largest margins on Arabic and English.
Taken together, these results show that
(a) our best approach is not only robust across languages, but also across (syntactic) parsing tasks, and
(b) the unfactorized approach may be well-suited to graph parsing tasks, which is in line with the results of \citet{dozat-manning-2018-simpler}.

\section{Discussion and Conclusion}
\label{sec:conclusion}
In this paper, we have performed a detailed and principled analysis on a variety of decisions arising during dependency parser design.
\textbf{What works?} We have identified an architecture based on fine-tuned XLM-R embeddings and factorized scoring that lead to new state-of-the-art performance for 11 out of 12 diverse language in our study on basic UD parsing, and for 5 out of 7 lanugages for enhanced UD parsing.
\textbf{What doesn't?} Adding LSTM layers on top of the transformer leads to a decrease in accuracy in most cases. We have also shown that multi-task setups predicting UPOS and UFeats often degrade parsing performance.
\textbf{What is really necessary?} For current state-of-the-art UD parsers, we recommend making sure that the pre-trained language model covers the intended domain well. In addition, keeping a factorized approach is a good idea for tree parsing, while in graph parsing, a single scorer module may suffice.

In this paper, we have addressed a high- to medium-resource scenario, assuming that we know the application language of a parser and thus training a single parser per language.
Future work may address multilingual approaches such as the training setup used by UDify or the recently proposed UDapter \cite{ustun-etal-2020-udapter}, which aims at boosting performance of low-resource languages while keeping performance of high-resource languages high.
Furthermore, it would be interesting to see if our results about biaffine achitectures also hold for non-syntactic tasks that have recently been framed as dependency parsing tasks, such as Named Entity Recognition \cite{yu-etal-2020-named}, negation scope detection \citep{kurtz-etal-2020-end} or Semantic Role Labeling \cite{shi-etal-2020-semantic}.

To sum up, in this paper we have applied \enquote{Occam's razor} to graph-based dependency parsing.
We believe that the insights from our study will foster further research on dependency parsing and on framing other tasks as dependency parsing, taking our simplified but robustly performing STEPS parser as a starting point. 

\section*{Acknowledgments}
We thank Agnieszka Falenska, Heike Adel, Lukas Lange, Hendrik Schuff, Jannik Strötgen, as well as the anonymous reviewers for their valuable comments with regard to this work.

\bibliography{references}
\bibliographystyle{acl_natbib}

\end{document}